\documentclass[sn-mathphys,Numbered]{sn-jnl}

\usepackage{graphicx}%
\usepackage{multirow}%
\usepackage{amsmath,amssymb,amsfonts}%
\usepackage{amsthm}%
\usepackage{mathrsfs}%
\usepackage[title]{appendix}%
\usepackage{xcolor}%
\usepackage{textcomp}%
\usepackage{manyfoot}%
\usepackage{booktabs}%
\usepackage{algorithm}%
\usepackage{algorithmicx}%
\usepackage{algpseudocode}%
\usepackage{listings}%

\usepackage[font=small,labelfont=bf]{caption}
\usepackage[T1]{fontenc}
\usepackage[utf8]{inputenc}

\theoremstyle{thmstyleone}%
\theoremstyle{thmstyletwo}%

\theoremstyle{thmstylethree}%

\begin{document}

\title[SF-TMN for Surgical Phase Recognition]{SF-TMN: SlowFast Temporal Modeling Network for Surgical Phase Recognition}


\author*[1]{\fnm{Bokai} \sur{Zhang}}\email{zhangbokai1994@gmail.com}

\author[2]{\fnm{Mohammad Hasan} \sur{Sarhan}}\email{msarhan@its.jnj.com}
\equalcont{These authors contributed equally to this work.}

\author[3]{\fnm{Bharti} \sur{Goel}}\email{bhartigoel0812@gmail.com}
\equalcont{These authors contributed equally to this work.}

\author[1]{\fnm{Svetlana} \sur{Petculescu}}\email{sveta0704@gmail.com}

\author[1]{\fnm{Amer} \sur{Ghanem}}\email{ghanemar@gmail.com}

\affil*[1]{\orgname{Johnson \& Johnson MedTech}, \orgaddress{\street{1100 Olive Way, Suite 1100}, \city{Seattle}, \postcode{98101}, \state{WA}, \country{USA}}}

\affil[2]{\orgname{Johnson \& Johnson MedTech}, \orgaddress{\street{ Robert-Koch-Straße 1}, \city{Norderstedt}, \postcode{22851}, \state{Schleswig-Holstein}, \country{Germany}}}

\affil[3]{\orgname{Johnson \& Johnson MedTech}, \orgaddress{\street{5490 Great America Pkwy}, \city{Santa Clara}, \postcode{95054}, \state{CA}, \country{USA}}}




 \abstract{\textbf{Purpose:} Automatic surgical phase recognition is one of the key technologies to support Video-Based Assessment (VBA) systems for surgical education. Utilizing temporal information is crucial for surgical phase recognition, hence various recent approaches extract frame-level features to conduct full video temporal modeling.

 \textbf{Methods:} For better temporal modeling, we propose SlowFast Temporal Modeling Network (SF-TMN) for surgical phase recognition that can not only achieve frame-level full video temporal modeling but also achieve segment-level full video temporal modeling. We employ a feature extraction network, pre-trained on the target dataset, to extract features from video frames as the training data for SF-TMN. The Slow Path in SF-TMN utilizes all frame features for frame temporal modeling. The Fast Path in SF-TMN utilizes segment-level features summarized from frame features for segment temporal modeling. The proposed paradigm is flexible regarding the choice of temporal modeling networks.
 \\
 \textbf{Results:} We explore MS-TCN and ASFormer models as temporal modeling networks and experiment with multiple combination strategies for Slow and Fast Paths. We evaluate SF-TMN on Cholec80 surgical phase recognition task and demonstrate that SF-TMN can achieve state-of-the-art results on all considered metrics. SF-TMN with ASFormer backbone outperforms the state-of-the-art Not End-to-End(TCN) method by $2.6\%$ in accuracy and $7.4\%$ in the Jaccard score. We also evaluate SF-TMN on action segmentation datasets including 50salads, GTEA, and Breakfast, and achieve state-of-the-art results. 
 \\
 \textbf{Conclusion:} The improvement in the results shows that combining temporal information from both frame level and segment level by refining outputs with temporal refinement stages is beneficial for the temporal modeling of surgical phases.
 }
 
\keywords{Surgical Phase Recognition, SlowFast, Segment, Frame, Temporal Modeling, Action Segmentation}



\maketitle

\section{Introduction}\label{sec1}

Video-based assessment (VBA) utilizes video recordings of procedures to evaluate the performance of the surgeons as well as support surgical education\cite{feldman2020sages}. Automatic surgical phase recognition can not only support video-based surgical workflow analysis in VBA systems but also help surgeons document and locate their case recordings efficiently in their online surgical video library.

Early studies~\cite{twinanda2016endonet,zia2018surgical} utilize image classification networks to capture spatial information from video frames to achieve surgical phase recognition. As spatial modeling from video frames cannot provide enough information to recognize the action of the surgeons, temporal modeling has become more and more popular in recent studies~\cite{jin2021temporal,jin2022trans,zhang2022towards,kirtac2022surgical,demir2022deep,valderrama2022towards,goldbraikh2023bounded,konduri2023surgical,zang2023surgical,tao2023last,liu2023lovit}. Some studies focus on short video segment temporal modeling with the combination of CNNs and LSTMs~\cite{jin2017sv,jin2020multi} or 3DCNN~\cite{zhang2021surgical}. While some studies focus on full video temporal modeling, utilizing CNNs to extract features for the full videos, and MS-TCNs~\cite{czempiel2020tecno,fer2023artificial,zhang2021swnet} or Transformers~\cite{czempiel2021opera,zhang2022surgicala,zhang2022surgicalb} for full video temporal modeling.

In this paper, inspired by the SlowFast networks for video recognition \cite{feichtenhofer2019slowfast} that utilize two paths to modeling high frame rate and low frame rate inputs together, we propose SlowFast Temporal Modeling Network (SF-TMN) for surgical phase recognition based on surgical videos that utilize two paths to achieve frame-level full video temporal modeling and segment-level full video temporal modeling. Our SF-TMN is a two-stage method. In the first stage of training, we utilize ResNet50~\cite{he2016deep} as the image classification network for initial training with the video frames from the target dataset. Then we utilize ResNet50 to extract spatial features from the video frames. We use these full video features to train our SF-TMN for the second stage of training. The Slow Path in SF-TMN focuses on frame temporal modeling with frame-level features. The Fast Path in SF-TMN utilizes segment-level features summarized from frame features to achieve segment temporal modeling. Initial predictions are generated by merging the Slow Path features and Fast Path features and then refined in the refinement stages of our SF-TMN. 

Our contributions in this paper are listed as follows: (1) We extend the two paths modeling idea from SlowFast Networks~\cite{feichtenhofer2019slowfast} from video clips action recognition and classification domain to the video action segmentation domain that can be applied to long surgical video sequences. While the original SlowFast Networks takes different sampling rates of video frames from short video clips as the input for different paths, our SF-TMN takes full video features as the input and modeling on both frame-level features and segment-level features summarized from frame features. (2) Our proposed SF-TMN is flexible regarding the choice of temporal modeling networks. We explore MS-TCN~\cite{farha2019ms} and ASFormer~\cite{yi2021asformer} as different video action segmentation backbones. (3) We experiment with multiple combinations of strategies for Slow and Fast Paths for our SF-TMN. As SF-TMN produces predictions at every time point for long videos instead of producing one classification prediction for one short video clip in SlowFast Networks, we demonstrate that merging the Fast Path for segment modeling into the Slow Path for frame modeling is more suitable for our SF-TMN. (4) Our SF-TMN achieves new state-of-the-art results for Cholec80 surgical phase recognition \cite{twinanda2016endonet}. Our SF-TMN with ASFormer backbone achieves a $2.6\%$ improvement in accuracy compared to the state-of-the-art Not End-to-End(TCN) method \cite{yi2022not}. Our SF-TMN with ASFormer backbone outperforms the state-of-the-art Less is More(Timestamp) method \cite{wang2022less} by $2.9\%$ in precision and recall as well as $6.2\%$ in terms of the Jaccard score. (5) To demonstrate the robustness of our method, we evaluate our SF-TMN on the 50Salads \cite{stein2013combining}, the GTEA \cite{fathi2011learning}, and the Breakfast \cite{kuehne2014language} datasets which are non-medical related datasets that are widely used in the video action segmentation domain. Our SF-TMN with ASFormer backbone can also achieve state-of-the-art results on the 50Salads, the GTEA, and the Breakfast datasets.

\section{Method}
\begin{figure}[h]%
\centering
\includegraphics[width=1.0\textwidth]{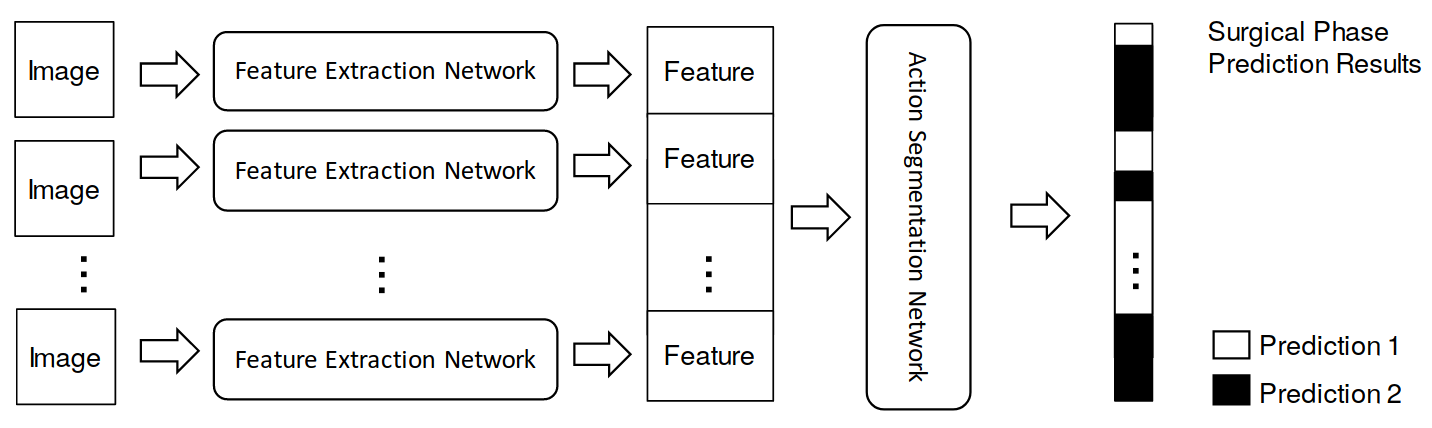}
\caption{The overview of our method }\label{fig1}
\end{figure}

The overview of our method is shown in Figure~\ref{fig1}. When modeling video clips with the action recognition networks like SlowFast Networks, the input sequences can be raw video frames. However, when modeling long video sequences, the input cannot be raw video frames at all time points as it will take tremendous GPU memory resources. To address this problem, we utilize a feature extraction network that is pretrained on the target dataset to extract features for every time point of the long video sequences. Next, we train an action segmentation network with full video features generated by concatenating the extracted features for surgical phase recognition. 

\subsection{Feature extraction network}
Feature extraction is performed on the frame level by extracting frame information using a ResNet50~\cite{he2016deep} model trained on the first 40 videos of the Cholec80 dataset~\cite{twinanda2016endonet}. The trained network is used as a feature extractor. The features of the frames contain spatial information and are used for training the proposed SF-TMN model. For the Cholec80 dataset, the videos are downsampled to 1 fps for feature extractions.

\subsection{Action segmentation network}
\begin{figure}[h]%
\centering
\includegraphics[width=1.0\textwidth]{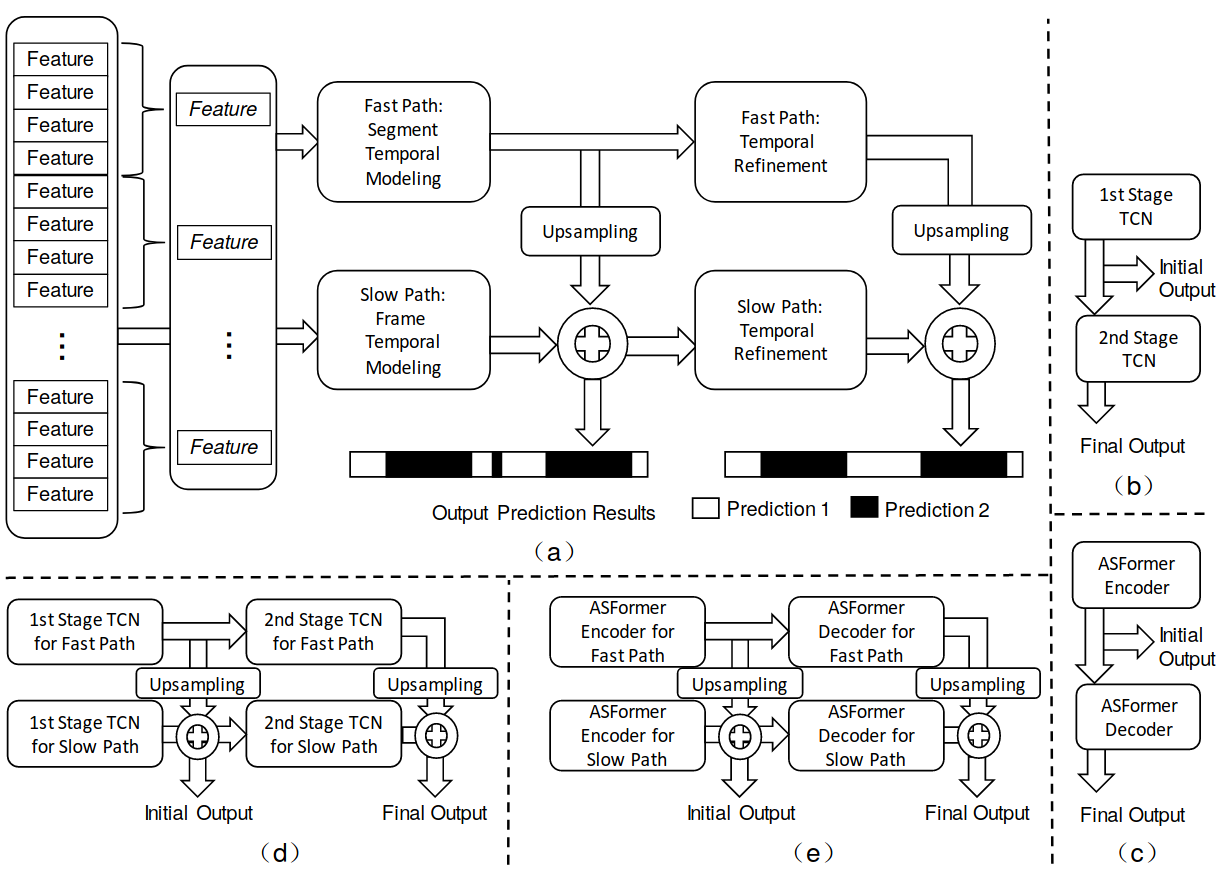}
\caption{Action segmentation network overview: (a) SF-TMN (b) MS-TCN (c) ASFormer (d) SF-TMN(MS-TCN) (e) SF-TMN(ASFormer)}\label{fig2}
\end{figure}
\subsubsection{MS-TCN}
Multi-Stage Temporal Convolutional Network (MS-TCN) proposed in~\cite{farha2019ms} is employed to capture temporal dependencies in video segments. For a video segment of length $T$ with frame sequence $(f_1, f_2, f_3,....f_T)$, an action $a_t$ is predicted at time $t$ in video $(0<=t<=T)$ by MS-TCN initially and later will be refined by the next stage of the network. 

\subsubsection{ASFormer}
Transformer for Action Segmentation (ASFormer)~\cite{yi2021asformer} is a model proposed to solve action segmentation problems with inspiration from transformer attention models~\cite{vaswani2017attention}. ASFormer follows an encoder-decoder architecture as shown in Figure~\ref{fig2} (c). The inputs to the encoder are the extracted features from a full video and the outputs are initial predictions as well as the features. The decoder then refines the initial predictions with previous output features. 
The encoder consists of multiple encoder blocks where each has a dilated temporal convolution and single-head self-attention layer. The decoder blocks consist of a dilated temporal convolution followed by a cross-attention layer that incorporates outputs from both the encoder and the previous layer. 

\subsubsection{SF-TMN}
Inspired by the SlowFast networks for video recognition \cite{feichtenhofer2019slowfast}, we design our SlowFast Temporal Modeling Network (SF-TMN) that consists of a Slow Path that focuses on video frame temporal modeling and a Fast Path that focuses on video segment temporal modeling. The inputs to the Slow Path are frame-level features and the inputs to the Fast Path are segment-level features downsampling from the frame-level features. Similar to other previous temporal modeling networks, our SF-TMN generates initial predictions in the temporal modeling stage and refines the initial predictions for our final output predictions. Multiple temporal refinement stages can be utilized to improve the output predictions further. One design of our SF-TMN is shown in Figure~\ref{fig2} (a). Figure~\ref{fig2} (a) consists of one temporal modeling stage and one temporal refinement stage for each path. We upsample the output features of the temporal modeling stage in the Fast Path and combine them with the output features of the temporal modeling stage in the Slow Path to generate the initial output predictions. The temporal refinement stage in the Slow Path utilizes the previous output predictions generated by the combined features while the temporal refinement stage in the Fast Path utilizes the output predictions from the previous stage of the Fast Path. The final output predictions are obtained by combining the output features from the Slow Path and the upsampled output features from the Fast Path. The combined features $F_{i}$ in the $i$th stage of our SF-TMN can be calculated by
\begin{equation}
F_{i} = w_{1,i} \times F_{s, i}  + w_{2,i} \times F_{f, i} 
\end{equation}
Where $F_{s, i}$ represents the features generated in the $i$th stage of the Slow Path. $F_{f, i}$ represents the upsampled features generated in the $i$th stage of the Fast Path. $w_{1,i}$ and $w_{2,i}$ are weighted parameters that are learned during the training.

Our SF-TMN can utilize different temporal networks as backbones. A two-stage SF-TMN with MS-TCN (Figure \ref{fig2} (b)) as the backbone is shown in Figure \ref{fig2} (d). A two-stage SF-TMN with ASFormer (Figure~\ref{fig2} (c)) as the backbone is shown in Figure~\ref{fig2} (e).

\begin{figure}[h]%
\centering
\includegraphics[width=1.0\textwidth]{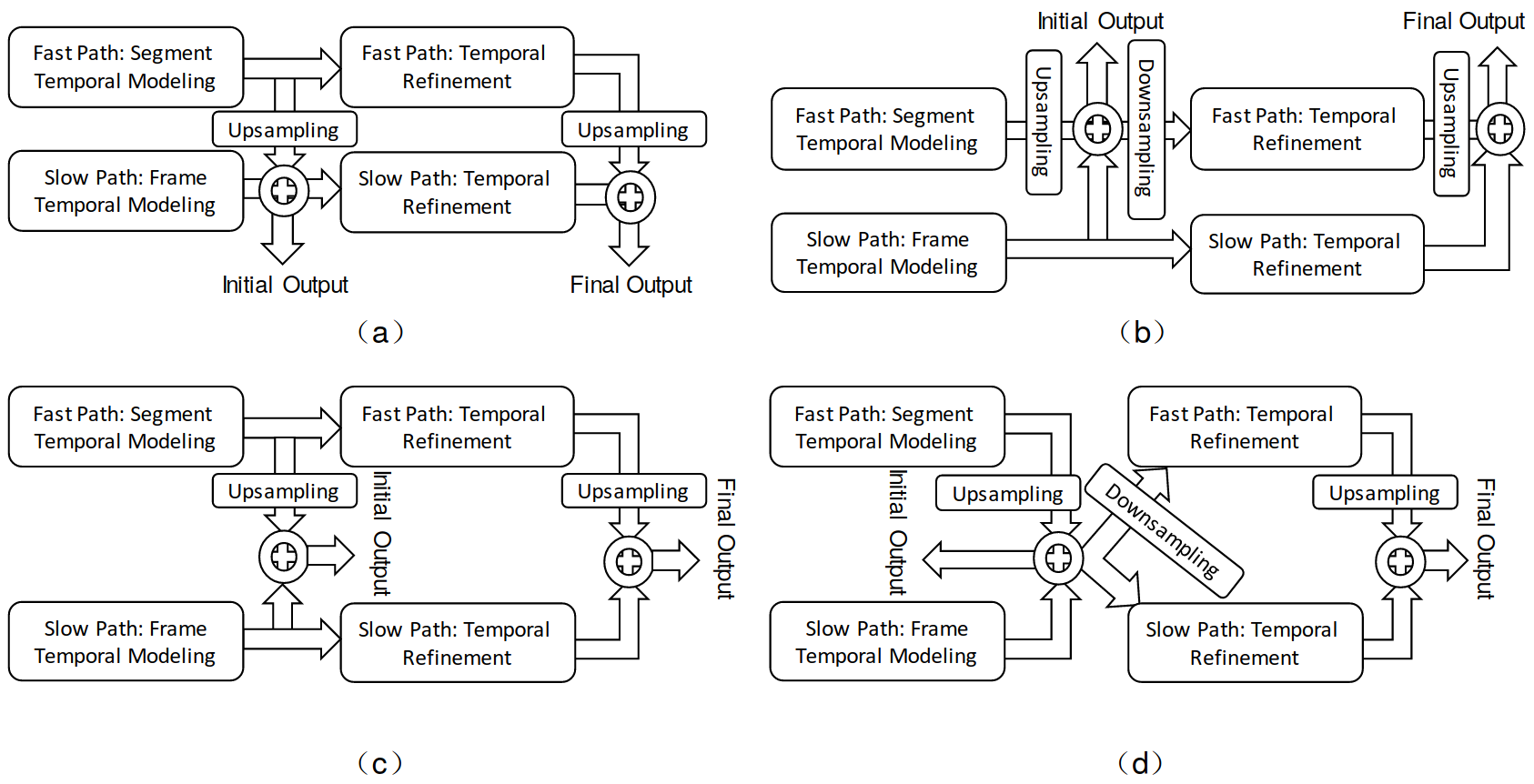}
\caption{Different designs for SF-TMN: (a) Slow Path temporal refinement utilizes the combined output (b) Fast Path temporal refinement utilizes the combined output (c) Neither Slow Path temporal refinement nor Fast Path temporal refinement utilizes the combined output (d) Both Slow Path temporal refinement and Fast Path temporal refinement utilize the combined output}\label{fig3}
\end{figure}

Different designs of SF-TMN are demonstrated in Figure~\ref{fig3}. The configurations in Figure~\ref{fig3} are two-stage SF-TMNs. The SF-TMN in Figure~\ref{fig3}(a) utilizes the combined output in the Slow Path for temporal refinement. This is the same design as Figure~\ref{fig2}(a). The SF-TMN in Figure~\ref{fig3}(b) utilizes the combined output in the Fast Path for temporal refinement. The SF-TMN in Figure~\ref{fig3}(c) does not utilize the combined output for temporal refinement. The SF-TMN in Figure~\ref{fig3}(d) utilizes the combined output for temporal refinement in both the Slow Path and the Fast Path. When utilizing the combined output for temporal refinement, different backbones utilize different combined data following the original design of the backbone networks. For SF-TMN with MS-TCN as the backbone, combined output predictions are utilized. For SF-TMN with ASFormer as the backbone, combined output predictions and combined output features are utilized. 

As the SlowFast networks are designed for short video sequence classification tasks, the high frame rate path (fine temporal resolution) is merged into the low frame rate path (coarse temporal resolution), so the SlowFast networks can generate one summarized final classification prediction for each short video sequence. Our SF-TMN, on the contrary, generates predictions for all time points in each long video sequence, so merging the segment path (coarse temporal resolution) into the frame path (fine temporal resolution) as shown in Figure~\ref{fig3}(a) seems to be a better design compared to Figure~\ref{fig3}(b). We will experiment with different designs of SF-TMN in the later section to validate this assumption. 

We utilized combined output predictions for loss calculation similar to MS-TCN and ASFormer. The total loss can be calculated by

\begin{equation}
L_{total} = L_{initial} + \sum_{N} L_{refined}
\end{equation}

where $L_{initial}$ is the loss calculated from the initial combined output predictions generated by the Fast Path for segment temporal modeling and the Slow Path for frame temporal modeling. $L_{refined}$ is the loss calculated from the combined output predictions generated by the Fast Path for temporal refinement and the Slow Path for temporal refinement, and $N$ is the total number of temporal refinement stages. Loss $L$ for temporal modeling stage $L_{initial}$ and temporal refinement stage $L_{refined}$ can be calculated by

\begin{equation}
L = L_{cls} + \lambda L_{smooth}
\end{equation}

where $L_{cls}$ is the cross-entropy loss, $\lambda$ is a weighted parameter, and $L_{smooth}$ is the smooth loss to reduce over-segmentation proposed in previous research \cite{farha2019ms,li2020ms}.

\section{Experiments}
\subsection{Dataset}
The Cholec80 dataset~\cite{twinanda2016endonet} is composed of 80 cholecystectomy surgery videos performed by 13 surgeons. The dataset includes annotations for the surgical phase and the surgical instrument presence. The 7 surgical phases are "P1: Preparation", "P2: Calot triangle dissection", "P3: Clipping and cutting", "P4: Gallbladder dissection", "P5: Gallbladder packaging", "P6: Cleaning and coagulation", and "P7: Gallbladder retraction". Following previous research,~\cite{twinanda2016endonet,wang2022less,yi2022not}, the first 40 videos were used for training, and the remaining 40 were used for testing.

The 50Salads dataset \cite{stein2013combining} contains 50 videos with 17 action classes and another two classes for "action start" and "action end". The 17 action classes are "cut tomato", "place tomato into bowl", "cut cheese", "place cheese into bowl", "cut lettuce", "place lettuce into bowl", "add salt", "add vinegar", "add oil", "add pepper", "mix dressing", "peel cucumber", "cut cucumber", "place cucumber into bowl", "add dressing", "mix ingredients" and "serve salad onto plate". For fair comparisons with other action segmentation methods, we utilized the extracted features provide by \cite{farha2019ms}. These features are extracted by Inflated 3D ConvNet (I3D) \cite{carreira2017quo}. The RGB stream features and the optical flow features are concatenated for each time point. Following previous research \cite{farha2019ms,li2020ms,yi2021asformer,wang2022cross,park2022maximization,liu2023diffusion}, five-fold cross-validation is performed on the 50Salads dataset and we use the same data splits.

The GTEA dataset \cite{fathi2011learning} contains 28 videos of 11 classes of daily activities in a kitchen. The 11 classes are "take", "open", "pour", "close", "shake", "scoop", "stir", "put", "fold", "spread", and "background". Following previous research \cite{farha2019ms,li2020ms,yi2021asformer,wang2022cross,park2022maximization,liu2023diffusion}, four-fold cross-validations are performed on the GTEA dataset and we use the same data splits. For fair comparisons with other action segmentation methods, we utilized the extracted features provide by \cite{farha2019ms}.

The Breakfast dataset \cite{kuehne2014language} contains 1712 videos with 48 classes related to cooking breakfasts in the third-person view. Following previous research \cite{farha2019ms,li2020ms,yi2021asformer,wang2022cross,park2022maximization,liu2023diffusion}, four-fold cross-validations are performed on the Breakfast dataset and we use the same data splits. For fair comparisons with other action segmentation methods, we utilized the extracted features provide by \cite{farha2019ms}.

\subsection{Evaluation metrics}
We utilize frame-level metrics \cite{twinanda2016endonet,wang2022less,funke2023metrics} and segmental metrics \cite{zhang2021swnet} to evaluate the performance of the model. Frame-level metrics evaluate the predictions of each frame while segmental metrics evaluate segments. We utilize segmental edit distance score~\cite{lea2016learning} and segmental $F1$ score~\cite{lea2017temporal} as segmental metrics. Intersection over Union (IOU) is used as the overlapping measure. To facilitate easier comparison, we also calculate $F1_{AVG}$ \cite{zhang2022surgicala} which is the average of the segmental $F1$ scores at $10\%$, $25\%$, and $50\%$ overlap thresholds that can be calculated by

\begin{equation}
    F1_{AVG}=\dfrac{1}{3} \times ({F1@10 + F1@25 + F1@50}) 
\end{equation}

\subsection{Implementation details}
We utilized the SGD optimizer with a learning rate of $1e^{-4}$ to train ResNet50, our feature extraction network, with cross-entropy loss. The weight decay was set to $1e^{-5}$. We set the batch size to 16 and the training epochs to 50. To achieve data augmentation, we resized the smaller side of the frames to 256 pixels and randomly cropped $224\times 224$ patches from the resized frames as the training samples. We also randomly selected 15\% of the training samples and randomly rotated them within 10 degrees to simulate camera rotation. 

For the Cholec80 dataset, We trained the MS-TCN, ASFormer, and SF-TMN with cross-entropy loss and smooth loss~\cite{farha2019ms}. We utilized the Adam optimizer with a learning rate of $1e^{-4}$. We set the batch size to 1 and the training epochs to 200. We set the total number of the dilated convolution layers at each stage, each encoder, and each decoder to 10. We set the number of feature maps to 64. We evaluated the effect of the number of stages in MS-TCN and the effect of the number of decoders in ASFormer in our Results section. The number of stages in MS-TCN was set to 4. We used one encoder and three decoders for ASFormer. We used one temporal modeling stage and three temporal refinement stages for each path of our SF-TMN. For fair comparisons between MS-TCN, ASFormer, and our SF-TMN, we utilized the same feature dataset for training which is generated with our ResNet50 feature extraction network.

For the 50Salads and GTEA datasets, we trained our SF-TMN with ASFormer backbone with cross-entropy loss and smooth loss~\cite{farha2019ms}. We utilized the Adam optimizer with a learning rate of $5e^{-4}$. We set the batch size to 1 and the training epochs to 200. We set the total number of the dilated convolution layers at each encoder and decoder to 10 and the number of feature maps to 64. We used one temporal modeling stage and three temporal refinement stages for each path of our SF-TMN. We use 32 features to generate one segment feature for the Fast Path. For the Breakfast dataset, we utilized the Adam optimizer with a learning rate of $1e^{-4}$. For fair comparisons between our method and previous studies, we utilized the extracted I3D features provide by \cite{farha2019ms} only and did not utilize extra multi-modal features \cite{li2022bridge, van2023aspnet} or hand pose features \cite{ishihara2022mcfm}.

\subsection{Results}

\subsubsection{Effect of the number of stages in MS-TCN}

We evaluate the effect of the number of stages in MS-TCN on the Cholec80 dataset. As shown in Table \ref{tab1r}, while different stages of TCNs perform similarly from the accuracy aspect,  the segmental edit distance score and the average segmental F1-score at the different thresholds improve. The 4-stage TCN outperforms the 2-stage TCN by approximately 14\% in terms of segmental edit distance score and the average segmental F1-score at the different thresholds. 

\subsubsection{Effect of the number of decoders in ASFormer}

We evaluate the effect of the number of decoders of ASFormer on the Cholec80 dataset. As shown in Table \ref{tab2r}, while ASFomers with different numbers of decoders perform similarly from the accuracy aspect,  the segmental edit distance score and the average segmental F1-score at the different thresholds improve. ASFomer with three decoders outperforms ASFomer with one decoder by approximately 5\% in terms of segmental edit distance score and the average segmental F1-score at the different thresholds. 

\begin{minipage}[t]{0.45\textwidth}
\centering
\fontsize{8}{10}\selectfont
\captionof{table}{Effect of the number of stages in MS-TCN on the Cholec80 dataset.}\label{tab1r}
\begin{tabular}[b]{llll}
\toprule
Stage & Accuracy & Edit & F1$_{AVG}$ \\
\midrule
Two &91.38    &68.10    &69.53 \\
Three &91.65   &80.41    &78.14 \\
Four &91.64    &82.40    &83.41 \\
\bottomrule 
\end{tabular}
\end{minipage}
\hfill
\begin{minipage}[t]{0.45\textwidth}
\centering
\fontsize{8}{10}\selectfont
\captionof{table}{Effect of the number of decoders in ASFormer on the Cholec80 dataset.}\label{tab2r}
\begin{tabular}[b]{llll}
\toprule
Decoder & Accuracy & Edit & F1$_{AVG}$ \\
\midrule
One &92.78    &79.05    &81.26 \\
Two &93.09   &82.37    &84.53 \\
Three &93.20    &84.47    &86.61 \\
\bottomrule 
\end{tabular}
\end{minipage}

\subsubsection{Effects of segment length} For our evaluation, we experimented with varying segment lengths for the Fast Path. Table~\ref{tab3} shows the performance scores with varying lengths of segments ranging from 4 frames per segment to 64 frames per segment. For the Cholec80 dataset, we extracted features with videos downsampled to 1 fps. As a result, the lengths of video segments are ranging from 4s to 64s. SF-TMN(MS-TCN) achieves the best performance with a segment length of 32 in all three performance matrices including accuracy, segmental edit distance score, and the average segmental F1-score at the different thresholds. Our SF-TMN(MS-TCN) with a segment length of 32 outperforms MS-TCN by approximately 1\% in accuracy, approximately 4\% in segmental edit distance score, and approximately 4.5\% in the average segmental F1-score.
For SF-TMN(ASFormer), a segment length of 32 also gives the best accuracy (94.72\%) overall with a comparable segmental edit distance score and a comparable average segmental F1-score. Our SF-TMN(ASFormer) with a segment length of 32 outperforms ASFormer by approximately 1.5\% in accuracy and the average segmental F1-score, approximately 2.5\% in terms of segmental edit distance score.

\begin{table}[h]
\centering
\caption{Effect of the segment length for the Fast Path in the SF-TMN on the Cholec80 dataset.}\label{tab3}
\begin{tabular}{lllll}
\hline
Method Name &  Segment Length & Accuracy & Edit & F1$_{AVG}$\\
\hline
{MS-TCN} & NA & 91.64    &82.40    &83.41 \\
\hline
{SF-TMN(MS-TCN)} & 4 & 91.28 & 81.74 & 84.60 \\
                & 8 & 91.25 & 83.46 & 85.75 \\
                & 16 & 92.52 & 85.61 & 87.93 \\
                & 32 & \textbf{92.54} & \textbf{86.56} & \textbf{88.07} \\
                & 64 & 91.96 & 84.41 & 86.62 \\
\hline
{ASFormer} & NA & 93.20    &84.47    &86.61 \\
\hline
{SF-TMN(ASFormer)} & 16 & 94.32 & \textbf{87.38} & \textbf{89.30} \\
                & 32 & \textbf{94.72} & 87.21 &  88.23 \\
\hline
\end{tabular}
\end{table}

\subsubsection{Performance of different designs for SF-TMN}
Table~\ref{tab4} demonstrates the performance of SF-TMN(MS-TCN) for different designs shown in Fig~\ref{fig3}(a)-(d). Note that three refinement stages are utilized for the results presented in Table~\ref{tab4} while method Fig~\ref{fig3}(a)-(d) only contains one refinement stage for easier demonstration purposes. Three temporal refinement stages are sequentially connected. The Slow Path and the Fast Path in each refinement stage are connected in the same way in each design. The design utilizing the Slow Path for temporal refinement of the combined outputs performed best with significant performance improvement over other designs in segmental metrics. Our SF-TMN(MS-TCN) achieves $92.54\%$ for the overall accuracy, $86.56\%$ for the segmental edit distance score, and $88.07\%$ for the average segmental F1-score. This validated that merging the segment path (coarse temporal resolution) into the frame path (fine temporal resolution) as shown in Figure~\ref{fig3}(a) is a better design compared to other design choices we proposed.

\begin{table}[h]
\centering
\caption{Different designs of SF-TMN(MS-TCN) with three refinement stages on the Cholec80 dataset.}\label{tab4}
\begin{tabular}{lllll}
\hline
Method Figure (One refinement stage) &  Combined Output & Accuracy & Edit & F1$_{AVG}$\\
\hline
Figure \ref{fig3}(a) & Slow Path & \textbf{92.54} & \textbf{86.56} & \textbf{88.07} \\
Figure \ref{fig3}(b) & Fast Path & 92.03 & 76.50 &  78.92 \\
Figure \ref{fig3}(c) & No Path & 91.23 &  70.33 & 73.14 \\
Figure \ref{fig3}(d) & All Paths & 92.32 & 79.84 & 82.83 \\

\hline
\end{tabular}
\end{table}

\subsubsection{Different designs of pooling for SF-TMN(ASFormer)}
We evaluate different designs of pooling for SF-TMN(ASFormer) on the Cholec80 dataset as shown in Table \ref{tab3r}. Results show that utilizing Max Pooling to generate segment features outperforms utilizing Average Pooling to generate segment features in terms of accuracy and the segmental edit distance score.

\begin{table}[h]
\centering
\caption{Different designs of pooling for SF-TMN(ASFormer) on the Cholec80 dataset.}\label{tab3r}
\begin{tabular}{llll}
\hline
Pooling Method & Accuracy & Edit & F1$_{AVG}$\\
\hline
Max Pooling &  \textbf{94.72} & \textbf{87.21} &  88.23 \\
Average Pooling &  94.49 & 86.23 &  \textbf{88.89} \\
Power-average Pooling &  94.08 &  80.85 & 84.68 \\

\hline
\end{tabular}
\end{table}

\subsubsection{Comparison with State-of-the-Art on Cholec80}
This section presents the video-by-video performance comparisons of our methods with other state-of-the-art methods on the Cholec80 dataset. Table~\ref{tab5} shows the results in detail. Table~\ref{tab5} clearly depicts that our approach outperforms the other state-of-the-art methods. Our SF-TMN with ASFormer backbone achieves a $2.6\%$ improvement in accuracy, a $5.9\%$ improvement in recall, and a $7.4\%$ improvement in Jaccard score compared to the state-of-the-art Not End-to-End(TCN) method \cite{yi2022not}. Our SF-TMN with ASFormer backbone outperforms the state-of-the-art Less is More(Timestamp) \cite{wang2022less} method by $3.5\%$ in accuracy, $2.9\%$ in precision and recall as well as $6.2\%$ in terms of the Jaccard score.

Our ResNet SF-TMN (ASFormer) outperforms ResNet ASFormer~\cite{yi2021asformer,zhang2022surgicala} by approximately $1\%$ for accuracy, precision, and recall. Our ResNet SF-TMN(ASFormer) outperforms ResNet ASFormer by approximately $1.5\%$ in terms of Jaccard score. Our ResNet SF-TMN(MS-TCN) outperforms ResNet MS-TCN~\cite{farha2019ms} by approximately $1\%$ for accuracy, recall, and Jaccard score. This demonstrates our SF-TMN can utilize different backbone networks and achieve better performance compared to the original backbone networks.

\begin{table}[h]
\centering
\caption{Video-by-video accuracy, precision, recall, and jaccard for different methods on Cholec80 dataset (mean $\pm$ std. \%)}\label{tab5}
\begin{tabular}{lllll}
\hline
Method Name &  Accuracy & Precision & Recall & Jaccard\\
\hline

PhaseNet \cite{twinanda2016endonet}& 89.1$\pm$5.4 & 82.5$\pm$9.8 & 86.6$\pm$4.5 & $-$ \\
EndoNet \cite{twinanda2016endonet}& 92.0$\pm$1.4 & 84.8$\pm$9.1 & 88.3$\pm$5.5 & $-$ \\
EndoNet+LSTM \cite{twinanda2017vision}  & 92.2$\pm$7.7 & 88.1$\pm$6.5 & 89.0$\pm$5.7 & $-$ \\
Less is More(Timestamp) \cite{wang2022less} & 91.9$\pm$5.6 & 89.5$\pm$4.4 & 90.5$\pm$5.9 & 79.9$\pm$8.5 \\
Not End-to-End(TCN) \cite{yi2022not} & 92.8$\pm$5.0 & $-$ & 87.5$\pm$8.3 & 78.7$\pm$9.4 \\
\hline
ResNet MS-TCN \cite{farha2019ms}  & 92.88$\pm$6.15 & 92.22$\pm$4.16 & 89.76$\pm$7.23 & 82.34$\pm$6.25\\
ResNet SF-TMN(MS-TCN) & 93.58$\pm$4.70 & 92.18$\pm$4.54 & 90.66$\pm$6.52 & 83.03$\pm$6.33 \\
ResNet ASFormer \cite{yi2021asformer,zhang2022surgicala} & 94.25$\pm$5.17 & 91.70$\pm$5.39 & 92.33$\pm$5.01 & 84.48$\pm$6.49 \\
ResNet SF-TMN(ASFormer) & \textbf{95.43$\pm$3.98} & \textbf{92.40$\pm$5.31} & \textbf{93.41$\pm$4.41} & \textbf{86.14$\pm$6.61} \\

\hline
\end{tabular}
\end{table}

\begin{figure}[h]%
\centering
\includegraphics[width=1.0\textwidth]{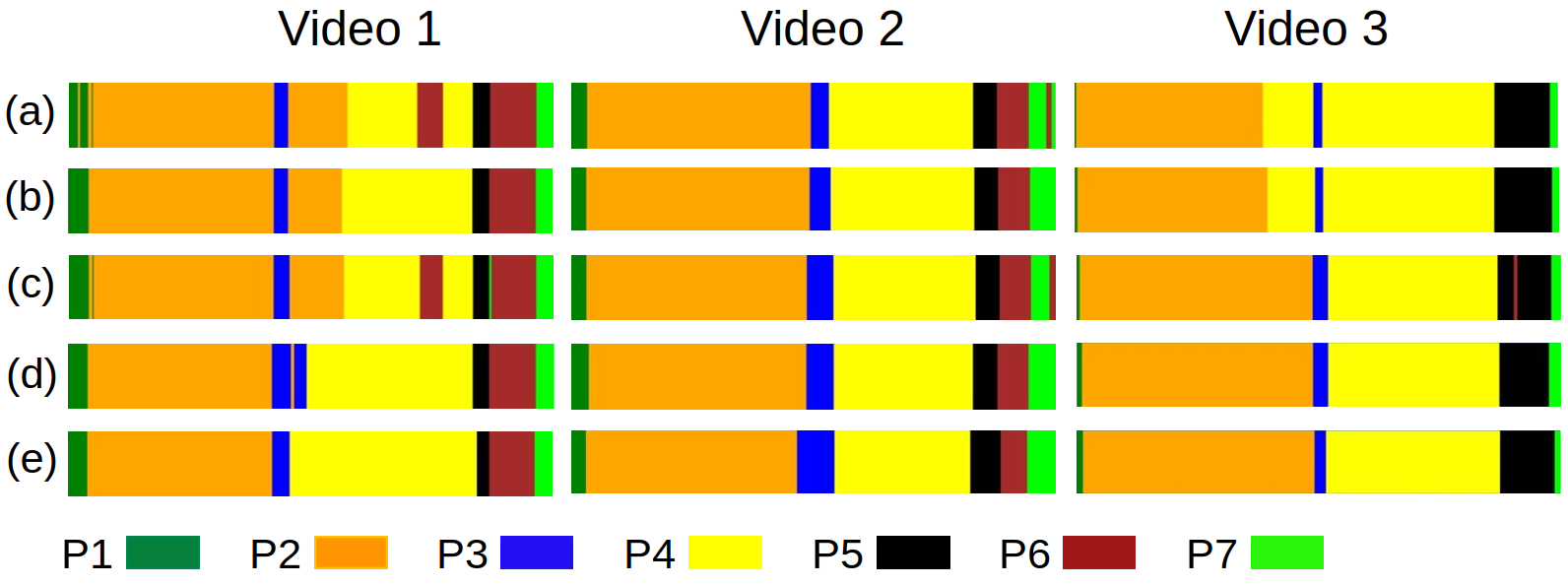}
\caption{Color-coded ribbon illustration for surgical
phase recognition: (a) MS-TCN (b) SF-TMN(MS-TCN) (c) ASFormer (d) SF-TMN(ASFormer) (e) Ground Truth}\label{fig4}
\end{figure}

We visualize predictions from state-of-the-art methods including ResNet MS-TCN and ResNet ASFormer along with our ResNet SF-TMN(ASFormer) and ResNet SF-TMN(MS-TCN) for three different test videos in Figure~\ref{fig4}. Predictions from SF-TMN(ASFormer) (row (d) in Figure~\ref{fig4}) are closest to ground truth segments (row (e) in Figure~\ref{fig4}) for all three videos. Whereas, SF-TMN(MS-TCN) (row (b) in Figure \ref{fig4}) is close to the ground truth except for some segment errors for P2.

\subsubsection{Comparison with State-of-the-Art on 50Salads, GTEA, and Breakfast}

To demonstrate our SF-TMN can also be applied to non-medical related videos in the action segmentation domain and can be trained with features generated by different feature extraction networks. We evaluate our SF-TMN with ASFormer backbone on the 50Salads, the GTEA, and the Breakfast datasets. The features for the 50Salads, the GTEA, and the Breakfast datasets are extracted by I3D and provided by \cite{farha2019ms}. 

As shown in Table \ref{tab6}, our SFTMN(ASFormer) outperforms ASFormer by 4.2\% in terms of accuracy, 4.8\% in terms of the segmental edit distance score, and 4\% to 6.9\% for the segmental F1-score at different thresholds on the 50Salads dataset. Our SFTMN(ASFormer) outperforms ASFormer by 3.3\% in terms of accuracy, 4.3\% in terms of the segmental edit distance score, and 1.8\% to 3.9\% for the segmental F1-score at different thresholds on the GTEA dataset. Our SFTMN(ASFormer) outperforms ASFormer by 3.5\% in terms of accuracy, 2\% in terms of the segmental edit distance score, and 2.7\% to 4.8\% for the segmental F1-score at different thresholds on the Breakfast dataset. Our SFTMN(ASFormer) also outperforms UVAST \cite{behrmann2022unified} by 2.4\% in terms of accuracy on the 50Salads dataset as well as outperforms DPRN \cite{park2022maximization} by 1\% in terms of accuracy on the GTEA dataset. Our SFTMN(ASFormer) outperforms TCTr \cite{aziere2022multistage} by 0.9\% in terms of the segmental edit distance score on the Breakfast dataset. Our SFTMN(ASFormer) outperforms TCTr \cite{aziere2022multistage} by 2.1\% to 3.7\% in terms of the segmental F1-score at different thresholds on the Breakfast dataset. These results demonstrate that our SFTMN(ASFormer) can outperform ASFormer as well as can achieve competitive performance compare with other SOTA methods on these non-medical related datasets. This solidifies the robustness of our method in addressing action segmentation challenges.

\begin{table}[h]
\centering
\caption{Comparison with other state-of-the-art methods on the 50Salads, the GTEA, and the Breakfast datasets}\label{tab6}
\setlength{\tabcolsep}{3pt}

\begin{tabular}{llllllllllll}
\hline

& 50Salads & & &  GTEA & & &  Breakfast & & &\\
\cmidrule{2-11}

Method Name &  F1@{10, 25, 50} & Edit & Acc.&  F1@{10, 25, 50} & Edit & Acc.&  F1@{10, 25, 50} & Edit & Acc.\\
\hline
MS-TCN \cite{farha2019ms}& 76.3 74.0 64.5 & 67.9 & 80.7 & 85.8 83.4 69.8 & 79.0 & 76.3 & 52.6 48.1 37.9 & 61.7 & 66.3\\
MS-TCN++ \cite{li2020ms} & 80.7 78.5 70.1 & 74.3 & 83.7 & 88.8 85.7 76.0 & 83.5 & 80.1 & 64.1 58.6 45.9 & 65.6 & 67.6\\
SSTDA \cite{chen2020action} & 83.0 81.5 73.8 & 75.8 & 83.2 & 90.0 89.1 78.0 & 86.2 & 79.8 & 75.0 69.1 55.2 & 73.7 & 70.2\\
BCN \cite{wang2020boundary}& 82.3 81.3 74.0 & 74.3 & 84.4 & 88.5 87.1 77.3 & 84.4 & 79.8 & 68.7 65.5 55.0 & 66.2 & 70.4\\
MTDA \cite{chen2020actionb} & 82.0 80.1 72.5 & 75.2 & 83.2 & 90.5 88.4 76.2 & 85.8 & 80.0 & 74.2 68.6 56.5 & 73.6 & 71.0\\
G2L \cite{gao2021global2local} & 80.3 78.0 69.8 & 73.4 & 82.2 & 89.9 87.3 75.8 & 84.6 & 78.5 & 74.9 69.0 55.2 & 73.3 & 70.7\\
HASR \cite{ahn2021refining} & 86.6 85.7 78.5 & 81.0 & 83.9 & 90.9 88.6 76.4 & 87.5 & 78.7 & 74.7 69.5 57.0 & 71.9 & 69.4\\
DTGRM \cite{wang2021temporal}& 79.1 75.9 66.1 & 72.0  & 80.0 & 87.8 86.6 72.9 & 83.0 & 77.6 & 68.7 61.9 46.6 & 68.9 & 68.3\\
ASRF \cite{ishikawa2021alleviating}& 84.9 83.5 77.3 & 79.3  & 84.5 & 89.4 87.8 79.8 & 83.7 & 77.3 & 74.3 68.9 56.1 & 72.4 & 67.6\\
UARL \cite{chen2022uncertainty} & 85.3 83.5 77.8 & 78.2 & 84.1 & 92.7 91.5 82.8 & 88.1 & 79.6 & 65.2 59.4 47.4 & 66.2 & 67.8\\
DPRN \cite{park2022maximization} & 87.8 86.3 79.4 & 82.0 & 87.2 & \textbf{92.9} \textbf{92.0} 82.9 & 90.9 & 82.0 & 75.6 70.5 57.6 & 75.1 & 71.7\\
TCTr \cite{aziere2022multistage} & 87.5 86.1 80.2 & 83.4 & 86.6 & 91.3 90.1 80.0 & 87.9 & 81.1 & 76.6 71.1 58.5 & 76.1 & \textbf{77.5}\\
FAM-MSDTN \cite{du2022dilated} & 86.2 84.4 77.9 & 79.9 & 86.4 & 91.6 90.9 80.9 & 88.3 & 80.7 & 78.5 72.9 60.2 & \textbf{77.5} & 74.8\\
UVAST \cite{behrmann2022unified} & \textbf{89.1} 87.6 81.7 & 83.9 & 87.4 & 92.7 91.3 81.0 & \textbf{92.1} & 80.2 & 76.9 71.5 58.0 & 77.1 & 69.7\\
\hline
ASFormer \cite{yi2021asformer}& 85.1 83.4 76.0 & 79.6  & 85.6 & 90.1 88.8 79.2 & 84.6 & 79.7 & 76.0 70.6 57.4 & 75.0 & 73.5 \\
SF-TMN(ASFormer) & \textbf{89.1} \textbf{88.0} \textbf{82.9} & \textbf{84.4} & \textbf{89.8} & 91.9 90.7 \textbf{83.1} & 88.9 & \textbf{83.0} & \textbf{78.7} \textbf{74.0} \textbf{62.2} & 77.0 & 77.0\\
\hline
\end{tabular}

\end{table}

\section{Applications}

\begin{figure}[h]%
\centering
\includegraphics[width=0.99\textwidth]{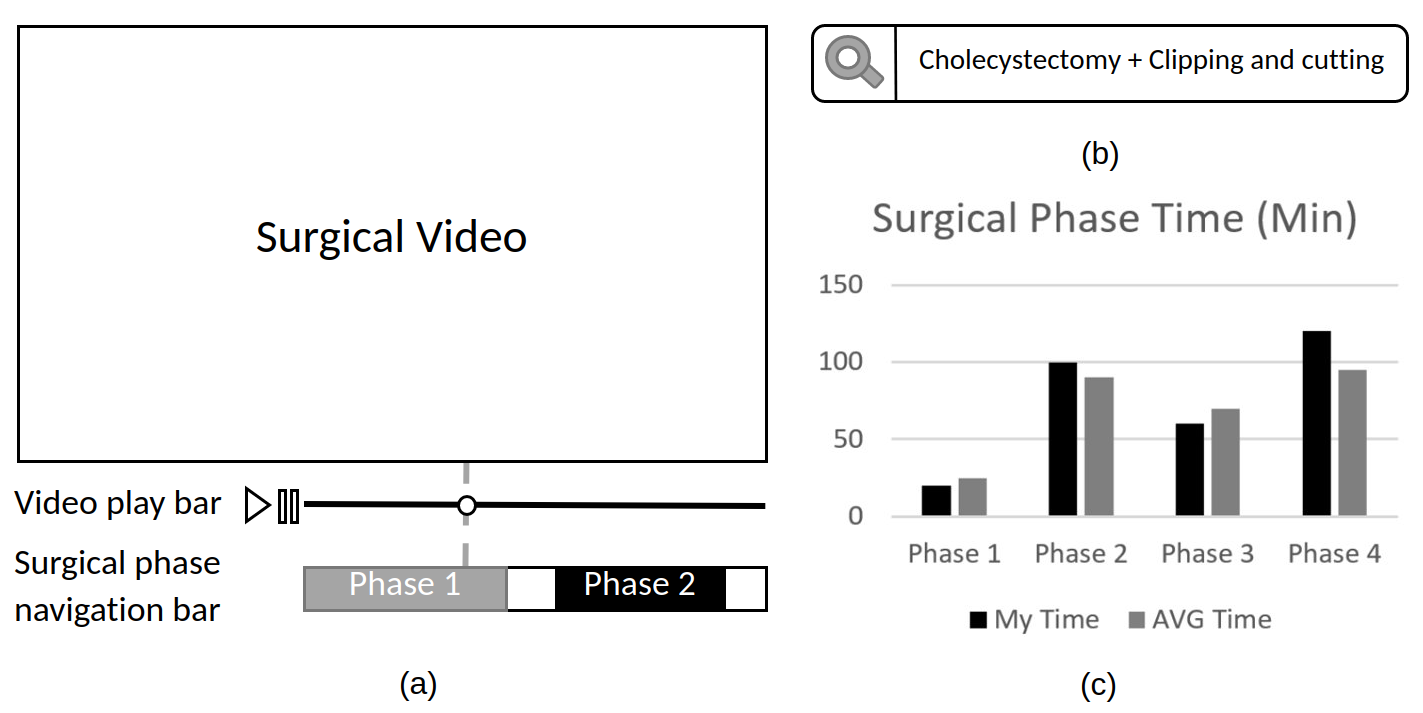}
\caption{Application examples for our surgical phase recognition model: (a) the surgical phase navigation bar (b) AI-based intelligent video search with keywords (c) Surgical phase time documentation and comparison.}\label{fig5}
\end{figure}

We share some application examples for our surgical phase recognition model in Figure \ref{fig5}. As shown in Figure \ref{fig5} (a), we can develop a surgical phase navigation bar to help surgeons to move to time periods of interest in the video in a more efficient manner for the online video library platform. As shown in Figure \ref{fig5} (b), we can utilize the surgical phase model to tag videos. Surgeons can utilize surgical phase names as keywords to search in the video library. As shown in Figure \ref{fig5} (c), the surgical phase models can help surgeons to document surgical phase time and compare with other expert surgeons to identify areas for improvement.

\section{Conclusion}

In this work, we propose SlowFast Temporal Modeling Network (SF-TMN). The proposed network utilizes multiple scales of temporal feature modeling and employs a multi-stage paradigm for temporal feature refinement. SF-TMN can support different temporal modeling backbones (e.g. MSTCN, ASFormer). We evaluate the model on Cholec80 surgical phase recognition and achieve state-of-the-art results on the accuracy, precision, recall, and Jaccard scores. This could be attributed to the fact that SF-TMN utilizes temporal information on two temporal scales (Slow and Fast) which allows the extraction of fine-grained information with the Slow Path while keeping the segment coarser information in the Fast Path. We also evaluate our SF-TMN on the 50Salads, the GTEA, and the Breakfast datasets to demonstrate the robustness of our methods for solving problems in the video action segmentation domain. A potential future direction is using more temporal scales to extract information on different layers of abstraction. Another direction is to experiment with this method in other surgeries and other surgical tasks.

\bibliography{sn-bibliography}

\section*{Ethics declarations}

\begin{itemize}
\item Conflict of interest

Bokai Zhang, Mohammad Hasan Sarhan, Bharti Goel, Amer Ghanem, and Svetlana Petculescu declare that they have no conflict of interest.

\item Ethics approval 

For this type of study, formal consent is not required. 

\item Informed consent

All experiments in this paper are conducted on open-source datasets. 

\end{itemize}

\end{document}